\documentclass[letterpaper, 10 pt, conference]{ieeeconf}  
\IEEEoverridecommandlockouts                              
\usepackage{verbatim}
\usepackage{graphics} 

\usepackage{amsthm}
\usepackage{amsmath,amssymb,stmaryrd,mathtools, amsfonts}
\usepackage{multirow}
\usepackage{hhline}
\usepackage[noend]{algpseudocode}
\PassOptionsToPackage{hyphens}{url}\usepackage{hyperref} 
\usepackage{dsfont}
\usepackage[ruled,linesnumbered]{algorithm2e}
\usepackage{multirow}
\usepackage[utf8]{inputenc}
\usepackage{import}
\usepackage{graphicx, graphics}
\usepackage{subcaption}
\usepackage{array}
\usepackage[keeplastbox]{flushend} 
\usepackage{color,xcolor}
\usepackage{soul}

\let\labelindent\relax
\usepackage{enumitem}
\definecolor{todo_c}{rgb}{1,0,0}

\usepackage{siunitx} 

\usepackage{svg}

\theoremstyle{remark}
\newtheorem{remark}{Remark}

\theoremstyle{definition}

\definecolor{gr}{RGB}{44, 160, 44}
\definecolor{or}{RGB}{255, 127, 14}
\definecolor{pur}{RGB}{128,0,128}

\title{\LARGE \bf
Collision Avoidance in Tightly-Constrained Environments without Coordination: a Hierarchical Control Approach
}

\author{Xu Shen$^\dag$, Edward L.\ Zhu$^\dag$, Yvonne R.\ St\"urz, and Francesco Borrelli%
\thanks{$\dag$Denotes equal contribution. The authors are with the Department of Mechanical Engineering, University of California at Berkeley, Berkeley, CA 94701 USA {\tt\small\{xu\_shen,edward.zhu\}@berkeley.edu}}
\thanks{This was supported by the European Union’s Horizon 2020 research and innovation programme under the Marie Sklodowska-Curie grant agreement No.\ 846421, the National Science Foundation under award No.\ 1931853, and the Office of Naval Research under grant No.\ N00014-18-1-2833}}

\begin{document}

\maketitle
\thispagestyle{empty}
\pagestyle{empty}



\begin{abstract}
We present a hierarchical control approach for maneuvering an autonomous vehicle (AV) in tightly-constrained environments where other moving AVs and/or human driven vehicles are present. 
A two-level hierarchy is proposed: a high-level data-driven strategy predictor and a lower-level model-based feedback controller.
The strategy predictor maps an encoding of a dynamic environment to a set of high-level strategies via a neural network. Depending on the selected strategy, a set of time-varying hyperplanes in the AV's position space is generated online and the corresponding halfspace constraints are included in a lower-level model-based receding horizon controller. These strategy-dependent  constraints drive the vehicle towards areas where it is likely to remain feasible.
Moreover, the predicted strategy also informs switching between a discrete set of policies, which allows for more conservative behavior when prediction confidence is low. We demonstrate the effectiveness of the proposed data-driven hierarchical control framework in a two-car collision avoidance scenario through simulations and experiments on a 1/10 scale autonomous car platform where the strategy-guided approach outperforms a model predictive control baseline in both cases.
\end{abstract}

\section{Introduction}
\label{sec:introduction}

Autonomous vehicles (AVs) have been the focus of significant research efforts in both academia and industry, where they have demonstrated notable potential in reducing the number of traffic incidents due to human-driver error. While autonomous driving tasks in well-structured environments, such as highway driving and lane changing~\cite{Pek2017} have largely been solved, more challenging tasks, such as driving in tightly-constrained environments, still poses open questions especially when other moving AVs and human driven vehicles are present.

The problem of collision avoidance, which is central to AV navigation, is nonconvex and NP-hard in general~\cite{Canny1987}. In tightly-constrained dynamic environments, additional challenges arise from these aspects: 1) nonlinear and non-holonomic vehicle dynamics, 2) non-convexity of environments and close proximity, and 3) change in obstacle configuration over time (motion of other human-driven or AVs).

Conventional search or sampling-based path planning methods, such as RRT, lattice planners, A$^*$ and their variants~\cite{Paden2016, Katrakazas2015, Dolgov2010} have been applied to the collision avoidance problem. However, these can suffer greatly from the curse of dimensionality and offer no guarantees of feasibility in dynamic environments. Reachability analysis~\cite{Vaskov2019, Shao2020} provides not-at-fault behavior with a parameterized set of trajectories, though these can be overly conservative in tightly-constrained spaces.

With the surge of machine learning and reinforcement learning (RL), there has been a strong interest in applying data-driven approaches in control. Among them, end-to-end planning~\cite{Schwarting2018, yang2018end} constructs a direct map from perception to control but lacks interpretability and safety guarantees. Deep RL has also been used for collision avoidance~\cite{Li2019, sun2019crowd}, but  typically requires discretization of the action space~\cite{Mirchevska18} or relaxation of the vehicle dynamics constraints.

Recently, optimization-based algorithms such as Model Predictive Control (MPC)~\cite{borrelli2017predictive} have received special attention due to their ability to include the exact dynamics and safety constraints in the formulation of the control problem. In~\cite{Zhang2017}, a novel optimization-based collision avoidance (OBCA) approach obtains a smooth reformulation of the collision avoidance constraints with exact vehicle geometry. 

In order to combine the advantages of different approaches, hierarchical frameworks for autonomous driving have been proposed in the literature. Such methods separate the problem into coarse path planning, maneuver choice, or parameter optimization at the higher strategic level, and trajectory planning, tracking, and control at the lower execution level~\cite{Schwarting2018, Paden2016, Katrakazas2015, Gonzalez2016, Walch2015}. As demonstrated in~\cite{vallon2020data}, MPC also displays great flexibility in leveraging data in a hierarchical manner, for example, by incorporating a learned ``strategy set'' to guide the receding horizon controller when navigating in an unknown environment.

In this work, we propose a hierarchical, data-driven, and strategy-guided control scheme to tackle the problem of motion planning in tight, dynamic environments. While our hierarchical framework is formulated for generic control of AVs in such environments, for simplicity, our presentation focuses on the specific scenario of collision avoidance between an ego vehicle (EV) and a target vehicle (TV). The more general case of pairwise collision avoidance between an EV and multiple TVs will be the subject of future investigation.
We assume that the vehicles do not use any form of motion coordination such as in \cite{Kneissl2020}.
This scenario is highly relevant, e.g., AV parking in a mixed autonomy setting, where collision avoidance and minimizing task duration are critical, and greatly depends on the selection of a good high-level strategy while maneuvering around other vehicles. In fact, a simple ``stop and wait'' strategy can be highly inefficient and might never complete the task in a crowded space.

\noindent Our contributions are twofold:
\begin{itemize}
    \item We propose a data-driven framework to construct a mapping from a high-dimensional environment encoding to a given set of high-level strategies. The latter is chosen such that it encodes prior knowledge of behavior in the corresponding scenario. Specifically, a neural network is trained offline with optimal collision-free trajectory rollouts, which are collected in a simulated environment. This mapping is used as a strategy predictor during online control.
    \item A strategy-guided time-varying MPC policy is formulated with exact vehicle and obstacle geometry to navigate a tightly-constrained dynamic environment. We refer to this policy as Strategy-Guided Optimization-Based Collision Avoidance (SG-OBCA). In addition to SG-OBCA, we also design a set of control policies which are selected based on the predicted strategy to maintain safety. The effectiveness of this control scheme is demonstrated through extensive numerical simulations and hardware experiments.
\end{itemize}


\noindent\textbf{Notation:} $\mathcal{B}_n(c,r)$ denotes the $l_2$-norm ball centered at $c \in \mathbb{R}^n$ with radius $r > 0$. The Minkowski sum of two sets $\mathcal{A}$ and $\mathcal{B}$ is denoted as $\mathcal{A} \oplus \mathcal{B} = \{a + b : a \in \mathcal{A}, b \in \mathcal{B}\}$. The minimum translation distance between two sets $\mathcal{A}$ and $\mathcal{B}$ is defined as $\text{dist}(\mathcal{A},\mathcal{B}) = \min_t \{\|t\| : (\mathcal{A}+t) \cap \mathcal{B} \neq \emptyset\}$ \cite{ericson2004real}. $\operatorname{vec}(\cdot)$ and $\operatorname{vec}(\cdot,\cdot)$ denote vectorization and concatenation operators to flatten a sequence of matrices into a single vector.
\section{Problem Formulation: Strategy-Guided Collision Avoidance} \label{sec:problem_formulation}

\begin{figure}[t!]
    \vspace{0.2cm}
    \centering
    \includegraphics[width=0.7\linewidth]{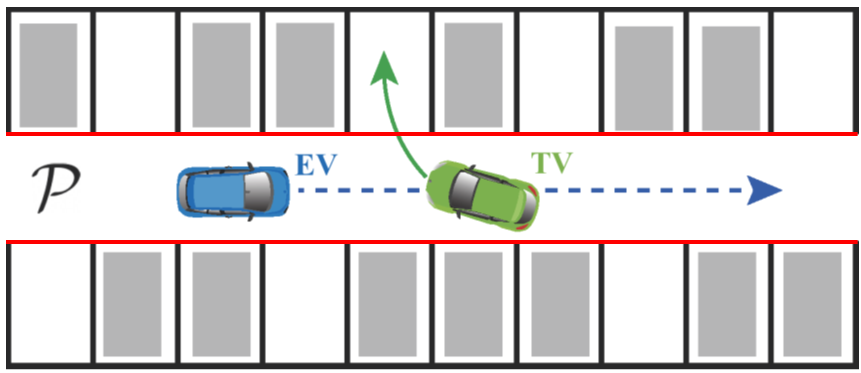}
    \caption{The EV tracks a reference \textcolor{blue}{(dashed blue)} in the local region $\mathcal{P}$ while the TV executes a parking maneuver \textcolor{gr}{(solid green)}. The lane boundaries are marked in \textcolor{red}{red}.}
    \label{fig:scenario}
    \vspace{-0.3cm}
\end{figure}

\begin{figure}[t!]
    \centering
    \includegraphics[width=0.9\linewidth]{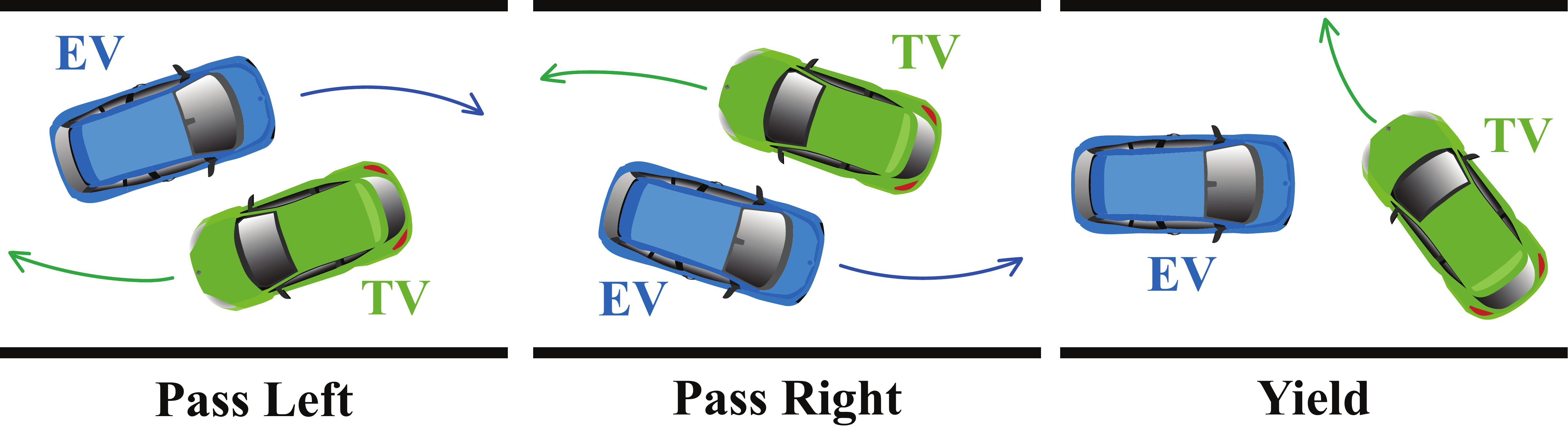}
    \caption{Available strategies in $\mathcal{S}$ for the EV navigation task}
    \label{fig:strategies}
    \vspace{-0.6cm}
\end{figure}

We consider the problem of two-vehicle collision avoidance in a parking lot between a human-driven TV and an autonomous EV. In particular, the scenario of interest occurs in a local region of the vehicles' position space $\mathcal{P} \subset \mathbb{R}^2$ and involves the TV executing a parking maneuver, while the EV seeks to navigate safely and efficiently through the narrow parking lot lane (Fig.~\ref{fig:scenario}).

In our scenario, the vehicles operate at low-speeds where tire slip and inertial effects can be ignored. We therefore model the vehicle dynamics using the kinematic bicycle model~\cite{Kong2015} given by
\begin{equation}
    \dot{z} := \left[
        \begin{matrix}
            \dot{x} \\ \dot{y} \\ \dot{\psi} \\ \dot{v}
        \end{matrix}
    \right] = \left[
        \begin{matrix}
            v\cos(\psi + \beta) \\ v\sin(\psi + \beta) \\ \frac{v}{l_r}\sin(\beta) \\ a
        \end{matrix}
    \right], \ u = \left[
        \begin{matrix}
            \delta_f \\ a
        \end{matrix}
    \right],
    \label{eq:kin_model_ct}
\end{equation}
where $\beta = \tan^{-1}\left( l_r\tan(\delta_f)/(l_f + l_r) \right)$ and the following nonlinear, time-invariant, discrete-time system with state $z_k \in \mathcal{Z} \subseteq \mathbb{R}^4$ and input $u_k \in \mathcal{U} \subseteq \mathbb{R}^2$ is obtained by discretizing \eqref{eq:kin_model_ct} with the 4-th order Runge–Kutta method
\begin{align}
    z_{k+1} = f(z_k, u_k), \label{eq:general_dynamics}
\end{align}

The geometry of the EV is modelled as a rotated and translated rectangle with length $L$ and width $W$, given as $\mathbb{B}(z_k) = R(\psi_k) \mathbb{B}_{0} + [x_k,y_k]^{\top}$ with $\mathbb{B}_{0} := \{ p\in \mathbb{R}^2: G p \leq g\}$, where $G = [I_2,-I_2]^{\top}, \ g = [L/2, W/2, L/2, W/2]^{\top}$, $p = (x,y) \in \mathbb{R}^2$ denotes the position states, and $R(\psi_k) \in SO(2)$ is the rotation matrix corresponding to the angle $\psi_k$.

The TV and lane boundaries comprise the dynamic environment and are parameterized by $\eta_k \in \mathbb{R}^o$. Specifically, we consider $M$ time-varying obstacles in the position space which can be each described as compact polyhedrons, i.e.
\begin{equation}
    \mathbb{O}^{(m)}_k := \{ p\in\mathbb{R}^2: A^{(m)}_k p \leq b^{(m)}_k\}, \ m = 1, \dots, M, \nonumber
\end{equation}
where $A^{(m)}_k \in \mathbb{R}^{q_m \times 2}$ and $b^{(m)}_k \in \mathbb{R}^{q_m}$ are known matrices at time $k$, and $q_m$ is the number of faces of the $m$-th polyhedron. The environment encoding $\eta_k$ can then be constructed as $\eta_k = \operatorname{vec}\left( \mathbb{O}^{(1:M)}_k \right) =  \operatorname{vec}\left(A^{(1:M)}_k, b^{(1:M)}_k\right)$. In the examples presented in this paper, we have $M=3$.

At every discrete time step $k$, we require that the EV remain collision-free with all $M$ obstacles. This is expressed as the constraint
\begin{equation}
    \begin{aligned}
        \mathrm{dist}\left(\mathbb{B}(z_{k}), \mathbb{O}^{(m)}_{k}\right) \geq d_{\mathrm{min}}, \ \forall m = 1,\dots, M,
    \end{aligned}
    \label{eq:dist_func}
\end{equation}
which enforces a minimum safety distance $d_{\mathrm{min}} > 0$ between the EV and obstacles. We assume that measurements and predictions of the time-varying obstacles are available over an $N$-step time horizon, i.e., at time step $k$, we have access to $\boldsymbol{\eta}_k = \{\eta_k, \eta_{k+1}, \dots, \eta_{k+N}\}$, and that there is no mismatch between the predictions and realized trajectories of the time-varying obstacles. The navigation task is then defined as a constrained tracking problem for the reference $\mathbf{z}_k^{\text{ref}}$.

Central to our approach is the selection a high-level strategy $s_k$ from a finite and discrete set $\mathcal{S}$, which encode prior knowledge of the behavioral modes available to the EV. In the case of our two-vehicle scenario, we use the following high-level strategies, which are illustrated in Fig.~\ref{fig:strategies}.
\begin{enumerate}[label=(\roman*)]
    \item \textbf{Pass Left}: The EV passes the TV from the left side;
    \item \textbf{Pass Right}: The EV passes the TV from the right side;
    \item \textbf{Yield}: The EV yields to the TV for safety.
\end{enumerate}

\begin{remark}
The strategy set $\mathcal{S}$ is provided by the designer and will require domain expertise in its construction. The problem of identifying salient strategies given a task is outside the scope of this work.
\end{remark}

{\setlength{\parindent}{0cm} \textit{Summary of Approach}}

\begin{figure}[t!]
    \vspace{0.2cm}
    \centering
    \includegraphics[width=\linewidth]{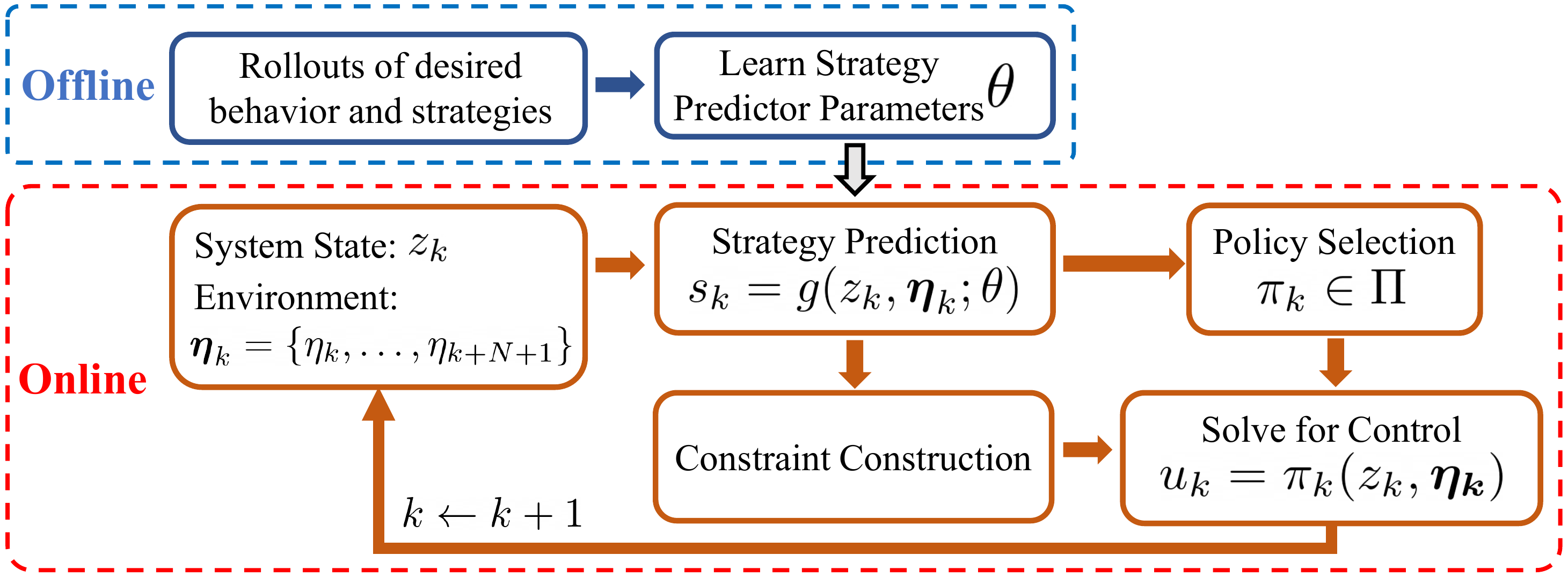}
    \vspace{-0.5cm}
    \caption{The proposed strategy-guided control scheme}
    \label{fig:flowchart}
    \vspace{-0.2cm}
\end{figure}

The constrained tracking task of interest is well-suited for MPC \cite{borrelli2017predictive}. However, control horizon lengths, for which real-time MPC is possible, may result in myopic behavior leading to collision in tightly-constrained environments.

Our proposed strategy-guided control scheme for collision avoidance (illustrated in Fig.~\ref{fig:flowchart}) attempts to address this issue by leveraging the high-level strategy as a surrogate for describing long-term behavior in the presence of other vehicles. In particular, we design a strategy predictor
\begin{equation}
    s_k = g(z_k,\boldsymbol{\eta}_k; \theta), \label{eq:general_strategy_selector}
\end{equation}
and train the parameters $\theta$ in~(\ref{eq:general_strategy_selector}) using a database of human-driven parking maneuvers in a simulated environment, and locally optimal solutions to the EV navigation tasks corresponding to each of the TV maneuvers.  We leverage the expressiveness of data-driven methods for strategy selection, as opposed to an end-to-end architecture directly for control as such grey-box approaches typically exhibit poor sample efficiency and require large amounts of data.

For the lower-level online strategy-guided control scheme, we construct three policies:
\begin{equation}
    \Pi^{\mathrm{SG}} = \{\pi^{\mathrm{SG-OBCA}}, \pi^{\mathrm{SC}}, \pi^{\mathrm{EB}}\}, \label{eq:strategy_policy_classes}
\end{equation}
where $\pi^{\mathrm{SG-OBCA}}$ is a nominal MPC controller based on OBCA \cite{Zhang2017,Zhang2019}, which uses the selected strategy to generate hyperplane constraints which help reduce myopic behavior and improve the ability of the EV to complete its navigation task. $\pi^{\mathrm{SC}}$ is a safety controller which can be activated due to infeasibility of SG-OBCA or ambiguity in the behavior of the TV. $\pi^{\mathrm{EB}}$ is an emergency brake controller, to be triggered in the event of impending collision. Conditions for switching between the three policies are discussed in Section~\ref{subsec:policy_selection}.

\section{Offline Learning of the Strategy Predictor}
\label{sec:offline}

\subsection{Data Collection}
\label{subsec:data_collection_offline}
\begin{figure}[t!]
     \centering
     \begin{subfigure}{0.68\linewidth}
         \includegraphics[width=\linewidth]{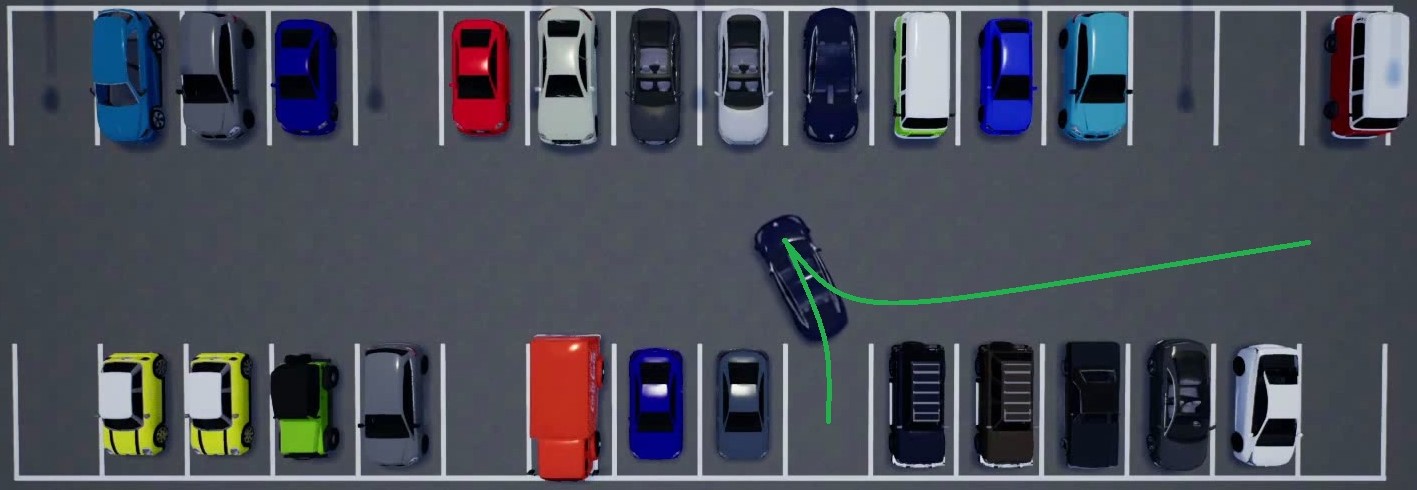}
         \caption{Parking Lot and Recorded Maneuver}
         \label{fig:carla_traj}
     \end{subfigure}
     \begin{subfigure}{0.305\linewidth}
         \includegraphics[width=\linewidth]{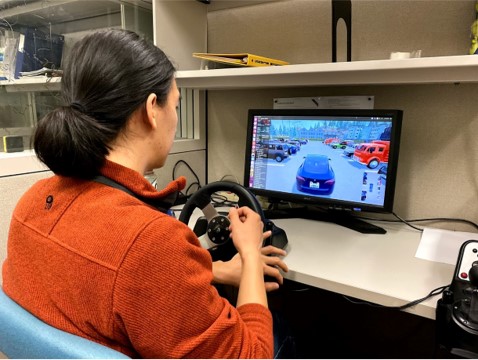}
         \caption{Subject Driving}
         \label{fig:carla_driver}
     \end{subfigure}
     \vspace{-0.1cm}
     \caption{Data Collection in CARLA}
     \label{fig:data_collection}
     \vspace{-0.5cm}
\end{figure}
We begin with raw data in the form of a set of human-driven parking maneuvers (Fig.~\ref{fig:carla_traj}). These were collected using the CARLA simulator~\cite{Dosovitskiy17} in a custom parking lot~\cite{shen2020parkpredict}, where the subject (driver) controls the brake, throttle, and steering of the vehicle (Fig.~\ref{fig:carla_driver}). For each parking trial, the parking lot is randomly populated with static vehicles and the subject was asked to park into different spots either in forward or reverse.

For the $i$-th recording of length $T^{[i]}$, the environment encoding is constructed as $\boldsymbol{\eta}^{[i]}_{0:T^{[i]}} = \left\{\operatorname{vec}\left( \mathbb{O}_k^{(1:M),[i]} \right)\right\}_{k=0}^{T^{[i]}}$, which contains the lane boundaries and the time-varying TV obstacles over the entire task. After obtaining $\boldsymbol{\eta}^{[i]}_{0:T^{[i]}}$, a finite-time optimal control problem, formulated using the OBCA algorithm~\cite{Zhang2017}, was used to solve for the locally optimal, collision-free EV trajectory $\{\mathbf{z}^{[i],*}, \mathbf{u}^{[i],*}\}$.

Lastly, given the strategy set $\mathcal{S} = $\{``Pass Left'', ``Pass Right'', ``Yield''\}, the strategy label $s^{[i]} \in \mathcal{S}$ for the $i$-th task can be generated automatically by looking at the relative configurations of the EV and TV over the EV solution trajectory, as illustrated in Fig.~\ref{fig:strategies}.

\subsection{Strategy Predictor Training}
Using the method described in Sec.~\ref{subsec:data_collection_offline}, we collect $K$ locally optimal TV-EV rollouts and their corresponding strategy labels offline. Now, given the horizon length $N$, the training dataset is constructed as $\mathcal{D} = \{(X^{[1]}_{0}, s^{[1]}), \dots, (X^{[1]}_{T^{[1]}-N}, s^{[1]}), (X^{[2]}_{0}, s^{[2]}), \dots, \allowbreak (X^{[K]}_{T^{[K]}-N}, s^{[K]})\}$.
We note that the strategy labels remain constant within the $i$-th recording since the behavioral mode for each TV-EV rollout is fixed. 
Each data point $X^{[i]}_{k}$ consists of the current state of the EV and the environment encoding along the $N$-step horizon, i.e., $ X^{[i]}_{k} = \operatorname{vec}\left( z^{[i],*}_{k}, \boldsymbol{\eta}_k^{[i]} \right), \ k \in \{0, \dots, T^{[i]}-N\}$. Using this dataset, we train a neural network (NN) for the predictor $g(\cdot)$ in \eqref{eq:general_strategy_selector} to predict the strategy $s_{k}$ from $z_k$ and $\boldsymbol{\eta}_k$ at every time step. The network architecture is composed of a fully connected hidden layer with 40 nodes, $\operatorname{tanh}$ activation function, and a $\operatorname{softmax}$ output layer for 3 strategies in $\mathcal{S}$. The objective function we minimize is the cross-entropy loss for classification tasks.

\section{Online Strategy Prediction and Control}
\label{sec:online_control}

At time step $k$ of the online task execution, the trained network with parameters $\theta^{*}$ returns confidence scores over the strategy set $\mathcal{S}$, i.e. $\hat{Y}_k = \operatorname{NN}(z_k, \boldsymbol{\eta}_k; \theta^{*}) \in \mathbb{R}^3$. The predicted strategy is chosen to be the one with the highest score, i.e., $\hat{s}_k = g(z_k, \boldsymbol{\eta}_k; \theta^{*}) = \arg \max_{s \in \mathcal{S}} \hat{Y}_k$. By leveraging the predicted scores $\hat{Y}_k$ and selected strategy $\hat{s}_k$, we proceed to construct time-varying hyperplanes and select the control policy $\pi_k$ from the set $\Pi^{\text{SG}}$.
\begin{figure}[t!]
    \vspace{0.2cm}
    \centering
    \begin{subfigure}{0.8\linewidth}
        \includegraphics[width=\linewidth]{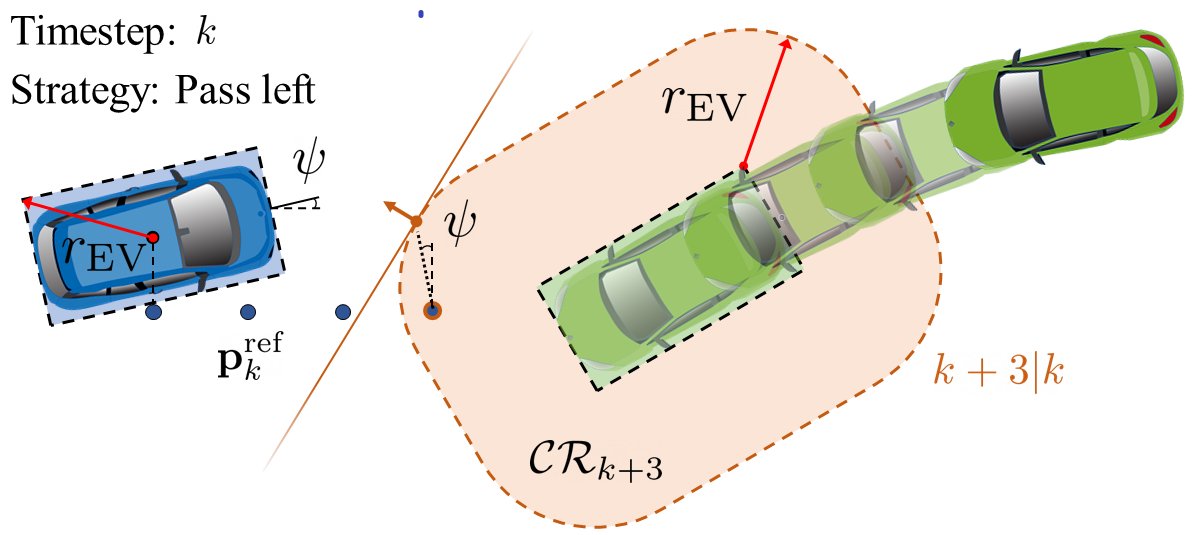}
        \caption{}
        \label{fig:const_gen_a}
    \end{subfigure}
    \begin{subfigure}{0.8\linewidth}
        \includegraphics[width=\linewidth]{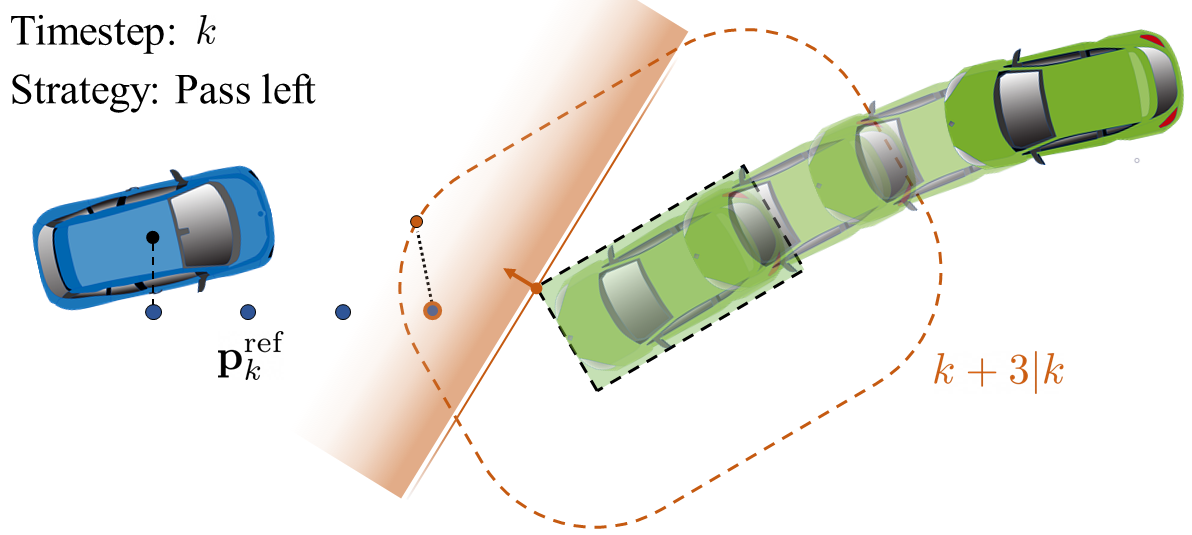}
        \caption{}
        \label{fig:const_gen_b}
    \end{subfigure}
    \begin{subfigure}{0.8\linewidth}
        \includegraphics[width=\linewidth]{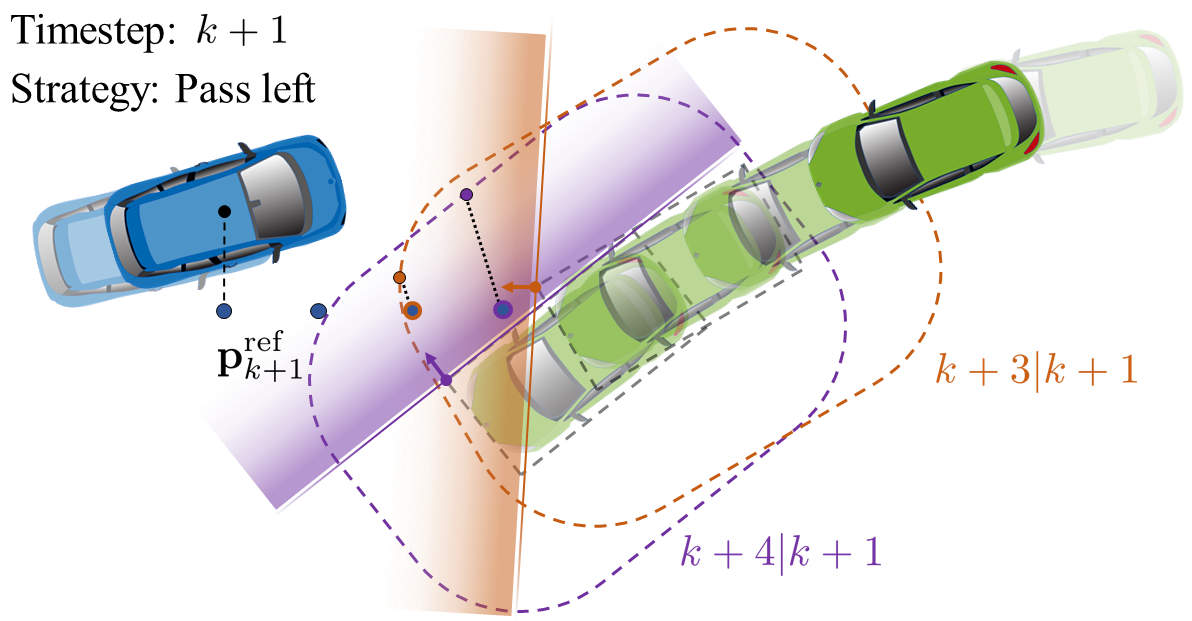}
        \caption{}
        \label{fig:const_gen_c}
    \end{subfigure}
    \vspace{-0.2cm}
    \caption{Illustration of the strategy-guided constraint generation procedure with $N=3$ and strategy ``Pass Left". a) The position reference of the EV at timestep $k+3$ falls within the critical region $\mathcal{CR}_{k+3}$. The reference point is projected to the boundary of the critical region in the positive body lateral direction and the tangent plane is found. b) The tangent plane is translated to be coincident with the TV boundary. c) Result of the constraint generation procedure at timestep $k+1$.}
    \label{fig:constraint_construction}
    \vspace{-0.6cm}
\end{figure}

\subsection{Constructing Constraints from Strategies}
\label{subsec:consrtraints_construction}

Similar to \cite{vallon2020data}, in this work, we view strategies as lower dimensional encodings of long term relative behavior between the controlled system and the dynamic environment. As such, we now describe how the chosen strategy at each timestep can be used to generate constraints for online control which help guide the EV into regions of space where it is more likely to successfully navigate around the TV.

For the TV occupying the region $\mathbb{O}_k^{\text{TV}}$ at timestep $k$, we define the critical region as:
\begin{align}
    \mathcal{CR}_k = \mathbb{O}_k^{\text{TV}} \oplus \mathcal{B}_2(0,r_{\text{EV}}), \nonumber
\end{align}
where $r_{\text{EV}} = \sqrt{L_{\text{EV}}^2 + W_{\text{EV}}^2}/2$ is the radius of the smallest disk which covers the extents of the EV as shown in Figure~\ref{fig:const_gen_a}. The critical region represents the area containing the TV where the collision avoidance constraints are close to or have become active.
 
Recall that we are interested in the tracking problem where we are given the $N$-step reference trajectory $\mathbf{z}_k^{\text{ref}} = \{z_{k|k}^{\text{ref}}, \dots, z_{k+N|k}^{\text{ref}}\}$ at time step $k$. In the scenario presented in Section~\ref{sec:problem_formulation}, this is simply a constant velocity trajectory which follows the centerline of the lane. Denote the position states of the reference trajectory as $\mathbf{p}_k^{\text{ref}}$.

The main idea behind our constraint generation procedure is to project positions along the reference trajectory, which fall within the critical region (i.e., $p_{k+t|k}^{\text{ref}} \in \mathcal{CR}_{k+t}$ for some $t \in \{0,\dots,N\}$), to the boundary of the critical region $\partial \mathcal{CR}_{k+t}$. The direction of projection is determined by the chosen strategy $\hat{s}_k$ of either ``Pass Left" or ``Pass Right". In particular, the two aforementioned strategies correspond with the directions $(-\sin \psi, \cos\psi)$ and $(\sin \psi, -\cos\psi)$ respectively, i.e., the positive and negative body lateral axis directions in the inertial frame. Since the boundary of the critical region is smooth by definition, we can find a unique tangent plane at the projection point with an outward facing normal vector $w_{k+t|k}(\eta_{k+t},\hat{s}_k)$ as can be seen in Figure~\ref{fig:const_gen_a}. The final step involves translating the tangent plane such that it is coincident with the boundary of $\mathbb{O}_k^{\text{TV}}$ as is shown in Figure~\ref{fig:const_gen_b}. This procedure results in a sequence of additional hyperplane constraints for time steps $k+t$ where $p_{k+t|k}^{\text{ref}} \in \mathcal{CR}_{k+t}$, $t \in \{0,\dots,N\}$. Denote the hyperplane parameters as $\phi(\eta_{k+t},\hat{s}_k) = -\text{vec}(w_{k+t|k}(\eta_{k+t},\hat{s}_k),b_{k+t|k}(\eta_{k+t},\hat{s}_k))$ and the augmented state $\bar{z}_{k+t|k} = \text{vec}(z_{k+t|k},1)$. The constraints can then be written as
\begin{align}
    \phi(\eta_{k+t},\hat{s}_k)^{\top}\bar{z}_{k+t|k}  \leq 0. \label{eq:strategy_hyperplane_constraint}
\end{align}
\begin{remark}
    While we may also apply the constraint generation procedure to the ``Yield" strategy, we choose to instead deal with it directly by activating a safety controller, which will be described in Sec.~\ref{subsec:policy_selection}.
\end{remark}

\subsection{Strategy-Guided Optimization-Based Collision Avoidance}

For the nominal MPC policy $\pi^{\mathrm{SG-OBCA}}$ in $\Pi^{\mathrm{SG}}$, we leverage the generated constraints to extend the OBCA algorithm~\cite{Zhang2017} which formulates the collision avoidance constraints using exact vehicle geometry to enable tight maneuvers in narrow spaces. By introducing the dual variables $\lambda^{(m)} \in \mathbb{R}^{q_m}, \mu^{(m)} \in \mathbb{R}^4$ associated with the $m$-th obstacle, we obtain a smooth reformulation of the collision avoidance constraint in \eqref{eq:dist_func}. The resulting NLP is written as follows:
\begin{subequations}
\begin{align}
    \min_{\mathbf{z, u}, \boldsymbol{\lambda}, \boldsymbol{\mu}} \ & \ \sum_{t=0}^{N} c(z_{k+t|k},u_{k+t|k},z_{k+t|k}^{\text{ref}}) \nonumber \\
    \text{s.t. } \ & z_{k+t+1|k} = f(z_{k+t|k},u_{k+t|k}), \  \scalebox{0.85}{$t=0,\dots,N-1$} \label{eq:obca_dynamics} \\
    & \ z_{k|k} = z_k, \label{eq:obca_init} \\
    & \scalebox{0.8}{$\left(A^{(m)}_{k+t|k} t(z_{k+t|k}) - b^{(m)}_{k+t|k}\right)^{\top} \lambda^{(m)}_{k+t|k} -g^{\top} \mu^{(m)}_{k+t|k} > d_\mathrm{min} $} \label{eq:dual_constr_1}\\
    & \ G^{\top} \mu^{(m)}_{k+t|k} + R(z_{k+t|k})^{\top} A^{(m) \top}_{k+t|k} \lambda^{(m)}_{k+t|k} = 0 \label{eq:dual_constr_2}\\
    & \ \left\|A^{(m) \top}_{k+t|k} \lambda^{(m)}_{k+t|k}\right\| \leq 1, \label{eq:dual_constr_3}\\
    & \ \lambda^{(m)}_{k+t|k} \geq 0, \mu^{(m)}_{k+t|k} \geq 0, \label{eq:dual_constr_4}\\
    & \ \phi(\eta_{k+t},\hat{s}_k)^{\top}\bar{z}_{k+t|k}  \leq 0, \label{eq:obca_strategy_constraint} \\
    & \ \forall t = 0, \dots, N, \quad m = 1, \dots, M, \nonumber
\end{align}
\label{eq:SG-OBCA}%
\end{subequations}
where $c(z_{k+t|k},u_{k+t|k},z_{k+t|k}^{\text{ref}}) =  \|z_{k+t|k}-z_{k+t|k}^{\text{ref}}\|_{Q_z}^2 + \|u_{k+t|k}\|_{Q_u}^2 + \|\Delta u_{k+t|k}\|_{Q_d}^2$ penalizes the deviation from the reference trajectory, input magnitude and rate, with $Q_z \succeq 0$, $Q_u \succ 0$, and $Q_d \succeq 0$, respectively. \eqref{eq:obca_dynamics} are the dynamics constraints over the control horizon, \eqref{eq:obca_init} sets the initial condition, \eqref{eq:dual_constr_1}-\eqref{eq:dual_constr_4} are the smooth and exact reformulations of \eqref{eq:dist_func} using Theorem~2 in~\cite{Zhang2017}, and \eqref{eq:obca_strategy_constraint} are the strategy-guided hyperplane constraints. We call the NLP in \eqref{eq:SG-OBCA}: SG-OBCA.

\begin{remark}
    Problem \eqref{eq:SG-OBCA} is non-convex, therefore providing a good initial guess to warm-start the NLP solver is crucial to maintain feasibility. For $\mathbf{z}$ and $\mathbf{u}$, it is straightforward to use the solution from the previous iteration as the initial guess. The initial guess for the dual variables $\boldsymbol{\lambda}$, and $\boldsymbol{\mu}$ are obtained with a similar method as described in~\cite{Zhang2019}.
\end{remark}

\subsection{Policy Selection} \label{subsec:policy_selection}


In our proposed strategy-guided control scheme we defined the set \eqref{eq:strategy_policy_classes}, which contains three classes of policies\footnote{Due to space constraints, the presented policy classes and their corresponding selection criteria are an abstraction of the actual implementation. Please visit \href{http://bit.ly/data-sg-control}{\texttt{\small bit.ly/data-sg-control}} for full details.}: SG-OBCA Control $\pi^{\mathrm{SG-OBCA}}$, Safety Control $\pi^{\mathrm{SC}}$, and Emergency Brake $\pi^{\mathrm{EB}}$. The formulation of each policy and the selection criteria are as follows.

\noindent\textbf{1) SG-OBCA Control:}
The SG-OBCA Control policy solves the problem \eqref{eq:SG-OBCA} at every time step $k$. 

\noindent\textit{Selection Criteria}:
\begin{enumerate}[label=(\roman*)]
    \item Problem (\ref{eq:SG-OBCA}) is solved successfully, \textbf{AND}
    \item $\hat{s}_k \in $ \{``Pass Left'', ``Pass Right''\} with high confidence, i.e. $\max_{s} \hat{Y}_k \geq \xi$, where $\xi$ is a user-defined threshold.
\end{enumerate}

\noindent\textbf{2) Safety Control:}
The Safety Control policy regulates the relative states between the EV and the TV into a safe control invariant set with respect to the two vehicles. In this work, we define this set to be the states with zero relative speed, such that the distance between EV and TV is preserved.

\noindent\textit{Selection Criteria}:
\begin{enumerate}[label=(\roman*)]
    \item Problem (\ref{eq:SG-OBCA}) is infeasible, \textbf{OR}
    \item $\hat{s}_k = $ ``Yield'' with $\max_{s} \hat{Y}_k \geq \xi$, \textbf{OR}
    \item $\max_{s} \hat{Y}_k < \xi$
\end{enumerate}

\noindent\textbf{3) Emergency Brake:}
We define the Emergency Brake policy in the context of impending system failure. When the Emergency Break policy is selected, it means that the autonomous agents cannot resolve the situation by either $\pi^{\mathrm{SG-OBCA}}$ or $\pi^{\mathrm{SC}}$ and human intervention is necessary.

\noindent\textit{Selection Criteria}: A collision is anticipated and cannot be resolved by either  $\pi^{\mathrm{SG-OBCA}}$ or $\pi^{\mathrm{SC}}$. 
\section{Simulation and Experimental Results} \label{sec:results}

\begin{figure}[t!]
    \vspace{0.2cm}
     \centering
     \begin{subfigure}{0.35\linewidth}
         \includegraphics[width=\linewidth]{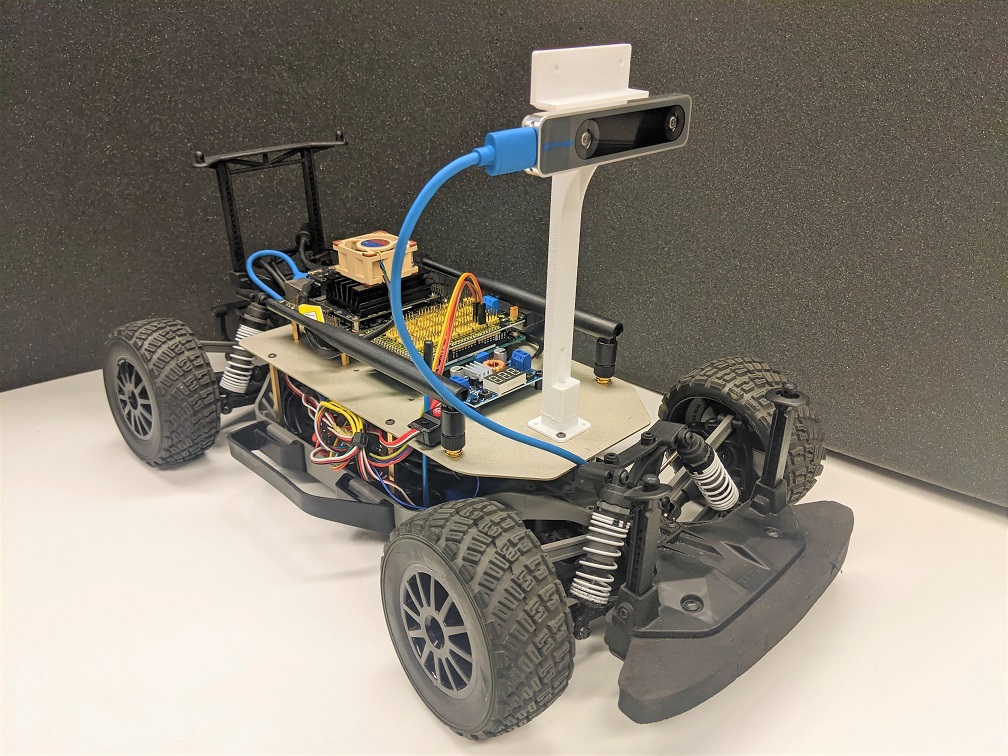}
         \caption{BARC}
         \label{fig:barc}
     \end{subfigure}
     \begin{subfigure}{0.495\linewidth}
         \includegraphics[width=\linewidth]{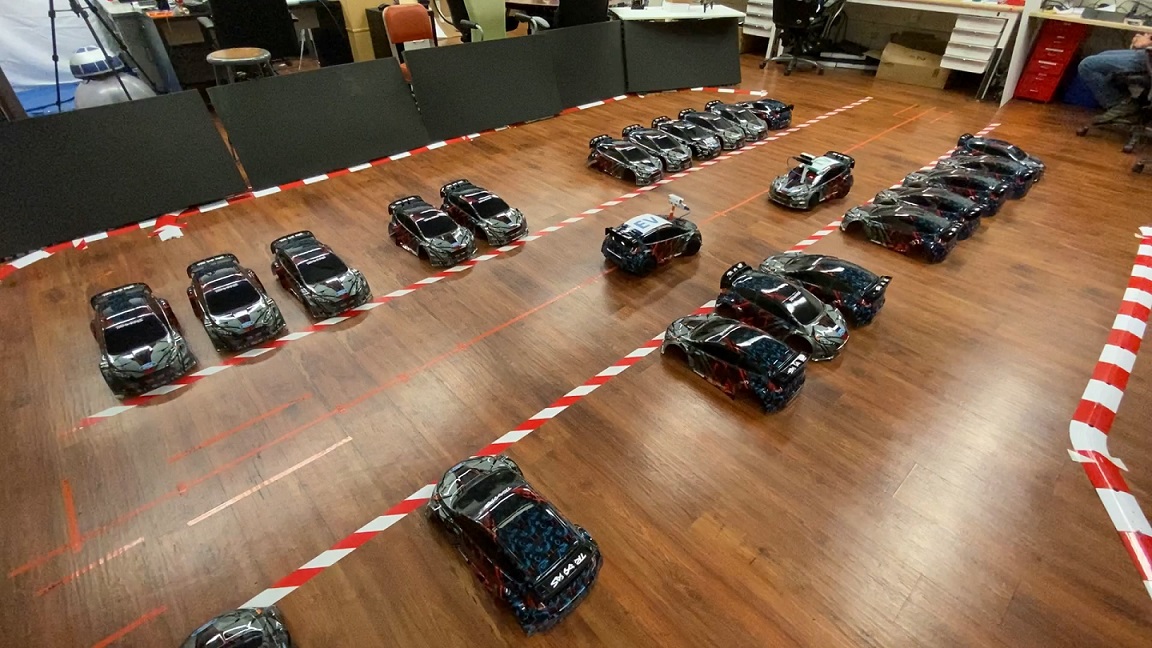}
         \caption{Parking Lot Environment}
         \label{fig:exp_lab}
     \end{subfigure}
     \vspace{-0.2cm}
     \caption{Experimental Setup}
     \label{fig:experiment}
     \vspace{-0.2cm}
\end{figure}
\begin{table}[t!]
    \centering
    \begin{tabular}{c|c|c|c}
        Control Method & Failure Rate & Avg. \# Iter & Median \# Iter\\
        \hline
        Baseline & 30\% & 213 & 220 \\
        \textbf{Strategy-Guided} & \textbf{6\%} & \textbf{270} & \textbf{227}
    \end{tabular}
    \caption{Failure Rate and Iterations to Task Completion}
    \label{tab:failure_rate_time}
    \vspace{-0.6cm}
\end{table}

In this section, we report results from simulations and experiments with the proposed strategy-guided control approach (SG) as described in Sec.~\ref{sec:online_control}. We compare the outcomes with a baseline control scheme (BL), where the $\pi^{\mathrm{SG-OBCA}}$ in \eqref{eq:strategy_policy_classes} is replaced with $\pi^{\mathrm{BL-OBCA}}$ which solves \eqref{eq:SG-OBCA} without the additional constraints \eqref{eq:obca_strategy_constraint}. Moreover, for the baseline control scheme, $\pi^{\mathrm{SC}}$ is selected only when $\pi^{\mathrm{BL-OBCA}}$ is infeasible. In both simulation and hardware experiments, we chose a horizon length of $N=20$ with a time step of $dt=0.1s$. The safety distance $d_{\text{min}}$ in \eqref{eq:dual_constr_1} is set to 0.01m and we set the confidence threshold to be 0.75. As described in Section~\ref{sec:offline}, we recorded a total of $486$ navigation tasks with TV maneuvers from which we generate $103, 553$ labeled examples for the training dataset $\mathcal{D}$. Details can be found at:  \href{http://bit.ly/data-sg-control}{\texttt{\small bit.ly/data-sg-control}}.

\begin{figure*}[t!]
    \vspace{0.2cm}
    \centering
    \begin{minipage}[c]{0.45\linewidth}
        \begin{center}
        \subfloat[]{
            \includegraphics[width=0.95\linewidth]{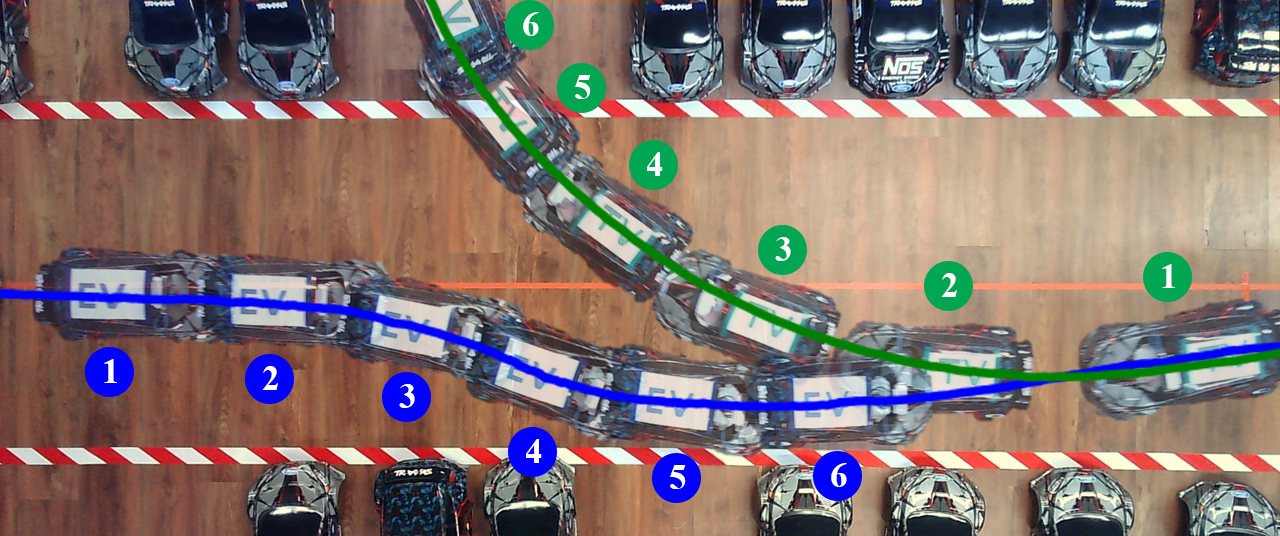} 
            \label{fig:2d_traj_6}
        } \\
        \subfloat[]{
            \includegraphics[width=0.95\linewidth]{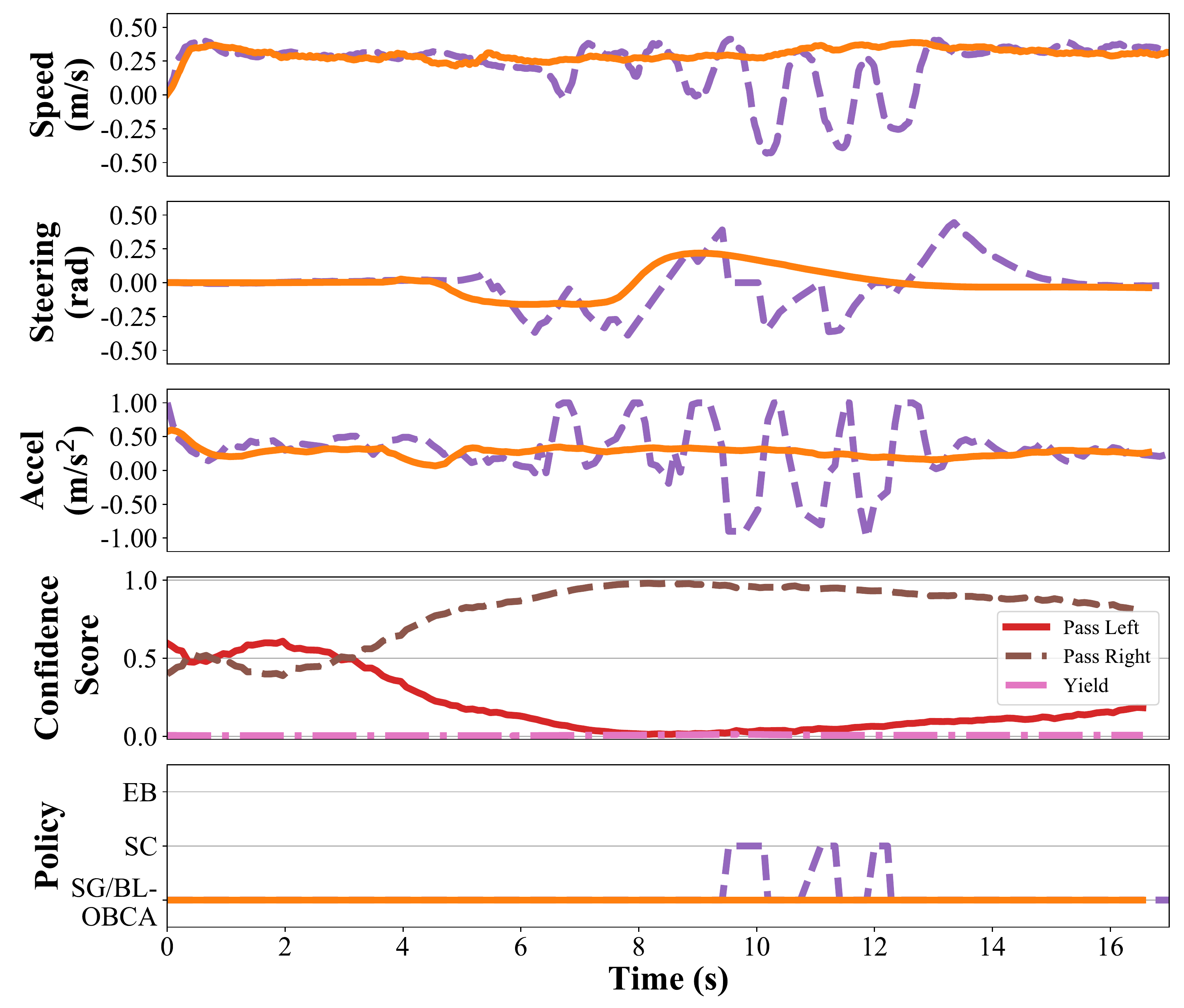} 
            \vspace{-0.3cm}
            \label{fig:speed_input_policy_6}
        }
        \end{center}
    \end{minipage}
    \begin{minipage}[c]{0.45\linewidth}
        \begin{center}
        \subfloat[]{
            \includegraphics[width=0.95\linewidth]{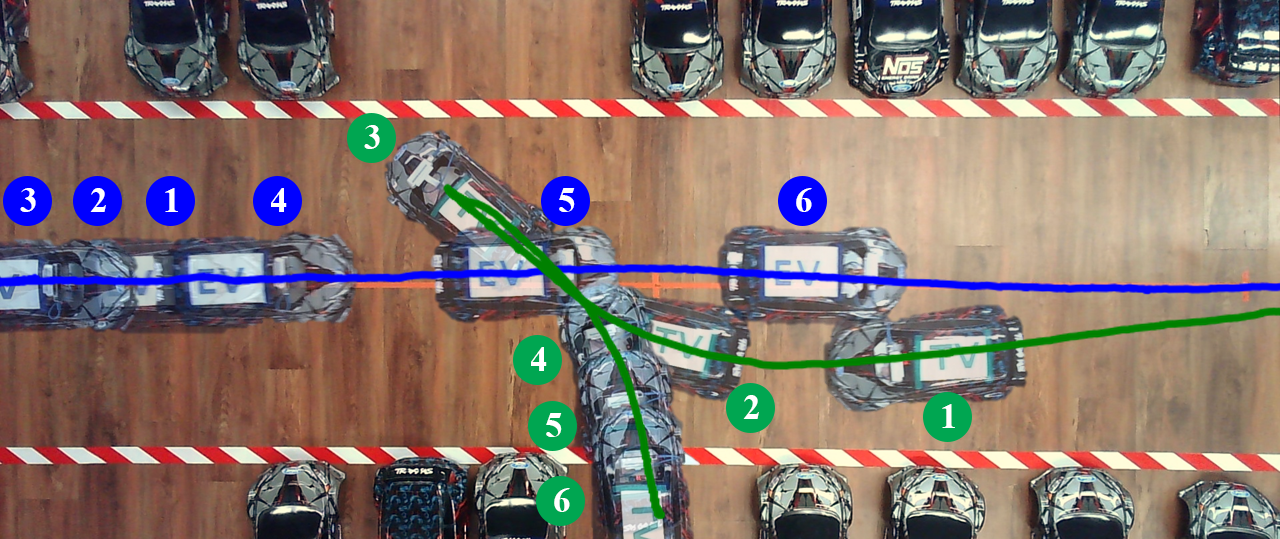} 
            \label{fig:2d_traj_10}
        } \\
        \subfloat[]{
            \includegraphics[width=0.95\linewidth]{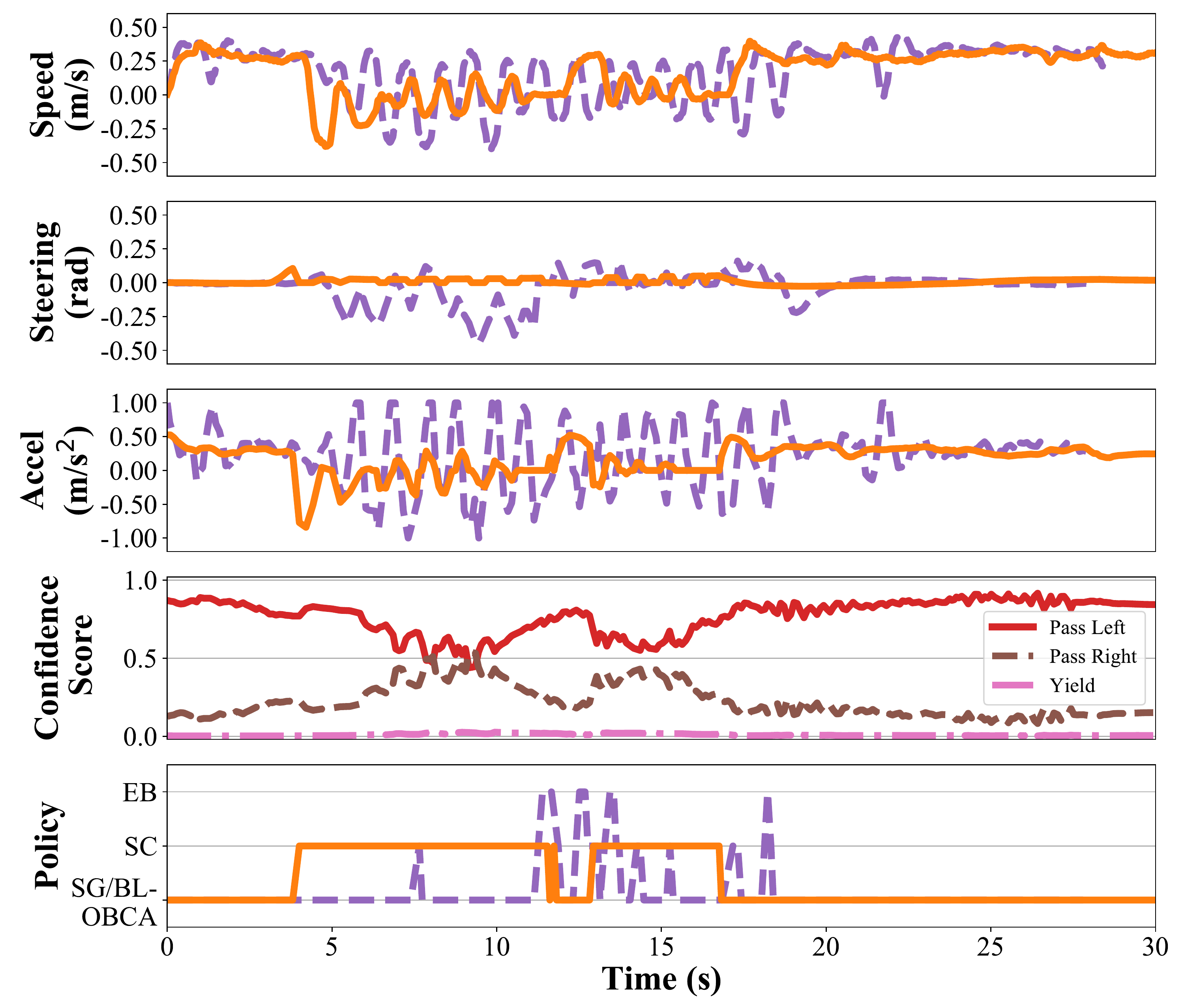} 
            \vspace{-0.3cm}
            \label{fig:speed_input_policy_10}
        }
        \end{center}
    \end{minipage}
    \vspace{-0.2cm}
    \caption{a), c) Trajectories of the \textcolor{blue}{EV (blue)} and \textcolor{gr}{TV (green)} under the strategy-guided control scheme. Frame numbers are marked with circles; b), d) (From top to bottom) Speed and input profile of the EV under \textcolor{or}{SG (solid orange)} and \textcolor{pur}{BL (dashed purple)}. Confidence of strategy prediction and policy selection for the strategy-guided controller.}
    \label{fig:exp_10}
    \vspace{-0.6cm}
\end{figure*}

All tasks are attempted by both SG and BL control schemes in a MATLAB simulation environment. Table~\ref{tab:failure_rate_time} reports the failure rate, where a collision occurs or the $\pi^{\mathrm{EB}}$ policy is selected, and the number of iterations required to complete the task. We can see that in simulation, the SG control scheme reduces the failure rate significantly, without sacrificing much in terms of task performance.

Both BL and SG approaches were implemented with the same cost matrices $Q_z = \text{diag}([1,1,1,10])$, $Q_u=\text{diag}([1,1])$, and $Q_d=\text{diag}([50,50])$, on the 1/10 scale open source Berkeley Autonomous Race Car (BARC) platform in a laboratory parking lot environment (Fig.~\ref{fig:experiment}). The cars are equipped with an Intel RealSense T265 tracking camera for localization and an NVIDIA Jetson Nano for onboard computation. We use FORCESPRO~\cite{FORCESNLP, FORCESPro} to compile the NLP solver, which achieves an average solution time of $30$ms for problem \eqref{eq:SG-OBCA}. $99.8\%$ of feasible solutions are returned in less than $100$ms. The strategy and policy selection, constraint generation, and NLP solver all run on a host laptop with an Intel i9 processor and 32 GB of RAM. Communication with the cars is done through the Robot Operating System (ROS). The human driven TV was represented by another car which tracked the recorded trajectories using nonlinear MPC. The MPC predictions at each time step are used to construct $\mathbb{O}_k^{\text{TV}}$ over the EV control horizon.

Results from the first navigation task are presented in Figs.~\ref{fig:2d_traj_6} and \ref{fig:speed_input_policy_6} where the TV parks into an empty spot in the top row. The predicted strategy and selected policy over the task are visualized in Fig.~\ref{fig:speed_input_policy_6}. It is clear that at the beginning of the task (before $t=4$s), the intention of the TV is ambiguous and the confidence in the ``Pass Left'' strategy is slightly higher because the TV veers to the right hand side of the EV. However, once the TV begins its parking maneuver, the ``Pass Right'' strategy is identified with high confidence score.  Under SG, the EV is able to leverage the hyperplane constraints generated via the identified strategy to preemptively begin its maneuver around the TV, thereby resulting in a smoother speed and input profile compared to BL as can be seen in Fig.~\ref{fig:speed_input_policy_6}. In contrast, BL performs a more aggressive and dangerous maneuver as it is only constrained by the safety distance $d_{\text{min}}$. This results in a fragile policy which can easily lead to constraint violation when the EV is in the critical region. Indeed, in Fig.~\ref{fig:speed_input_policy_6}, we observe that the BL triggers the Safety Control policy for several time steps due to infeasibility caused by the aggressive behavior.

Results from a second navigation task are presented in Figs.~\ref{fig:2d_traj_10} and \ref{fig:speed_input_policy_10} where the TV parks in reverse into an empty spot in the bottom row and a gear change is needed. The strategy and policy selections over the task are visualized in Fig.~\ref{fig:speed_input_policy_10}, where the SG control scheme selects the Safety policy $\pi^{\mathrm{SC}}$ at an early stage and maintains a safe distance for a considerable amount of time due to low confidence in the selected strategy. This is caused by the time the TV spends idling during the gear change. In contrast, the short-sighted baseline controller again exhibits overly-aggressive behavior by trying to pass the TV while it is idle. This eventually leads to activation of the policy $\pi^{\mathrm{EB}}$ and collision with the TV. Finally, we again see in Fig.~\ref{fig:speed_input_policy_10} that SG results in a smoother speed and input profile when compared to BL.
\vspace{-0.1cm}
\section{Discussion and Conclusion}
\label{sec:discussion}

In this work, we show that when compared to a baseline MPC approach, the proposed hierarchical data-driven control scheme significantly improves the success rate of a navigation task in a tightly-constrained dynamic environment. The design of a data-driven strategy predictor provides a structured and transparent approach to leveraging data in control, which is crucial to expanding the adoption of learning techniques in the control of safety critical systems.

The purpose of strategy-guided constraint generation and policy selection are to coerce the system towards regions of space where successful completion (in terms of recursive feasibility and stability) of the control problem is most likely. One interpretation of the strategy-guided constraints is that they are surrogates of the MPC terminal components, where given the selected strategy, we attempt to construct constraints which drive the state of the system \eqref{eq:general_dynamics} into a set which is contractive towards the reference $\mathbf{z}^{\text{ref}}$ while satisfying all constraints. Despite not affording the rigorous guarantees provided by an exact terminal cost function and terminal set, the value of our proposed surrogates lies in the fact that they can be straightforward to construct even when the exact formulations are intractable, which is typically the case for complex control tasks in dynamic environments.

In terms of the policy selection, predicting the strategy allows the system to ``prepare'' for an upcoming safety-critical encounter by selecting an appropriate control policy. In the context of this work, when faced with ambiguous behavior from the TV, the strategy-guided control approach can select the safety control policy, so that the system will only attempt the risky passing maneuver when it is confident enough to do so. In contrast, the baseline control scheme only reverts to the safety control when the nominal MPC becomes infeasible. This is usually at a time when the situation is already very challenging for the system to maintain safety.

Typical challenges due to uncertainty, brought about by localization errors, system delay, and model mismatch were encountered in the experimental setting. Therefore, future work will firstly focus on formulating a robust extension using both model-based and data-driven methods. Furthermore, applying the proposed approach to more complex environments with a higher number of dynamic objects or interactive/adversarial agents is of particular interest.

\newpage
\bibliographystyle{IEEEtran}

\bibliography{root.bib}

\end{document}